\documentclass[letterpaper, 10 pt, conference]{ieeeconf}

\IEEEoverridecommandlockouts
\usepackage{graphicx}
\usepackage{amsmath}
\usepackage{amssymb}
\usepackage{booktabs}
\usepackage{array}
\usepackage{tikz}
\usetikzlibrary{arrows.meta, positioning}
\usepackage[hidelinks]{hyperref}

\newcommand{\miou}{mIoU\textsubscript{comp}}

\title{\LARGE \bf
Technical Report for the ICRA 2026 GOOSE 2D Fine-Grained Semantic Segmentation Challenge: Pretraining-Diverse Ensemble of Foundation Vision Encoders for Robust Outdoor Scene Understanding}

\author{Boyan Wang$^{1}$, Yongxi Huang$^{1}$, Wenjing Li$^{1}$, Tianrui Hui$^{1}$,
Shaofei Huang$^{2}$, Nan Pu$^{1}$, and Zhun Zhong$^{1}$
\thanks{$^{1}$LION Lab, Hefei University of Technology, Hefei, China.}%
\thanks{$^{2}$University of Macau, Macau, China.}}

\begin{document}

\maketitle
\thispagestyle{empty}
\pagestyle{empty}

\begin{abstract}
This report presents our solution for the ICRA 2026 GOOSE 2D Fine-Grained Semantic Segmentation Challenge, which requires parsing unstructured outdoor scenes from four camera platforms into 56 fine-grained categories.
Our approach pairs foundation vision encoders (including DINOv3, SigLIP2, and InternImage) with a Mask2Former decoder, and trains them with a strong recipe including long training schedules, exponential moving average, a larger crop size, and multi-scale plus flip test-time augmentation.
The three encoders, chosen for their complementary pretraining objectives, are combined into a pretraining-diverse ensemble through per-class validation-IoU weighting.
Evaluated on the official GOOSE test set, our submission achieves 75.40\% composite mIoU and wins the second place of the challenge.
Our study further shows that the encoder's pretraining recipe, rather than its parameter count or the decoder design, is the dominant factor for accuracy on this benchmark.
\end{abstract}

\section{INTRODUCTION}
The GOOSE 2D Fine-Grained Semantic Segmentation Challenge benchmarks semantic segmentation for field robotics in unstructured outdoor environments.
The challenge contains images captured by four heterogeneous platforms, \textit{i.e.}, ALICE, MuCAR-3, Spot v1, and Spot v2, with resolutions ranging from $1280\times720$ to $2048\times1536$. Each pixel is assigned to one of 56 fine-grained categories, which are further grouped into 11 coarse super-categories.
The data distribution is highly long-tailed, with vegetation, terrain, and sky dominating the pixel budget, and it varies widely in viewpoint and illumination across platforms.

These properties make the perception backbone the central design decision.
Model design for this task can be improved along three axes, including the encoder that extracts image features, the decoder that predicts segmentation masks based on these features, and the training and inference recipe.
Because modern segmentation systems usually differ along all three axes at once, the relative importance of each is rarely isolated, and it is not obvious where the largest gains lie.

We study this question directly, fixing the decoder to a Mask2Former head~\cite{mask2former} and a common training budget while comparing pretrained encoders.
We find that large-scale pretrained foundation encoders, including those based on self-supervised, vision-language, or supervissed pretraining, such as DINOv3~\cite{dinov3}, SigLIP2~\cite{siglip2}, and InternImage~\cite{internimage}, transfer markedly better than encoders built around a novel architecture with standard ImageNet-22k supervision, such as Swin~\cite{swin} and ConvNeXt~\cite{convnext}, and that the gap is not explained by parameter count.

%
Building on this finding, our solution trains several strong foundation encoders with a simple recipe: a long training schedule, exponential moving average, a larger crop size, and multi-scale plus flip test-time augmentation.
We combine three encoders with complementary pretraining into a pretraining-diverse ensemble through per-class confidence weighting.
On the official GOOSE test set, our solution achieves 75.40\% composite mIoU and wins the second place.

Our contributions are summarized as follows:
\begin{itemize}
\item Through a controlled comparison under a fixed decoder and budget, we show that the encoder pretraining recipe is a major factor that explains performance differences beyond parameter count and decoder choices for fine-grained off-road semantic segmentation.
\item We present a simple training and inference recipe that substantially improves a single foundation encoder without architecture modification.
\item We combine three encoders with complementary pretraining into a per-class weighted ensemble as our final challenge submission, and analyze the role of pretraining diversity.
\end{itemize}

\section{RELATED WORK}
\textbf{Foundation vision encoders.}
Recent segmentation encoders are increasingly distinguished not only by their architectures but also by their pretraining recipes.
DINOv3~\cite{dinov3} scales self-supervised distillation to a 1.7B-image curated corpus; SigLIP2~\cite{siglip2} scales a sigmoid vision-language objective to web-scale image-text pairs; and InternImage~\cite{internimage} scales a deformable-convolution operator to billion-parameter regimes with large-scale supervision.
These pretraining-focused foundation encoders contrast with architecture-focused encoders such as Swin~\cite{swin} and ConvNeXt~\cite{convnext} which typically leverage ImageNet-1K or ImageNet-22K for pretraining.
%
ConvNeXtV2~\cite{convnextv2} moves architecture-focused ConvNets closer to foundation encoders by introducing masked autoencoding.
It is a common practice to transfer these pretrained knowledge to dense prediction tasks~\cite{huang2024modality,wang2023transferring}.
We treat ``foundation versus architecture-focused'' as the axis of interest and test it directly.

\textbf{Mask-classification decoders.}
Mask2Former~\cite{mask2former} casts segmentation as mask classification with a transformer decoder, and has become a strong alternative to per-pixel heads such as FPN head~\cite{fpn,huang2020ordnet}, Upernet~\cite{upernet} and the SegFormer MLP head~\cite{segformer} on benchmarks with many semantic categories and other related dense prediction tasks~\cite{huang2025revisiting}.
We adopt it as a fixed, strong decoder so that encoder effects are not confounded by decoder choice.

\textbf{GOOSE challenge solutions.}
The 2025 GOOSE 2D challenge winner~\cite{goose2025_2d} combined a RoPE-Swin backbone with color-shift correction and quantile-based label denoising.
Color augmentation has been a recurring theme in outdoor segmentation given the exposure and white-balance differences across platforms.
Motivated by these findings, we also examine color-shift correction and photometric distortion and find that neither helps when combined with a foundation visual encoder.

\section{METHOD}

\subsection{Task and metric}
The challenge ranks submissions by a composite metric that weights fine and coarse segmentation equally,
\begin{equation}
\miou = \tfrac{1}{2}\,\mathrm{mIoU}_{\text{fine}}
      + \tfrac{1}{2}\,\mathrm{mIoU}_{\text{coarse}},
\end{equation}
where $\mathrm{mIoU}_{\text{fine}}$ averages per-class IoU over the 56 evaluated classes and $\mathrm{mIoU}_{\text{coarse}}$ averages over their 11 super-categories.
IoU is accumulated over images from all four platforms before averaging, and eight classes are excluded from scoring.
Predictions are submitted at native input resolution as single-channel label maps.

\subsection{Architecture}
\begin{figure}[t]
\centering
\begin{tikzpicture}[
  node distance=4mm and 5mm,
  box/.style={draw, rounded corners, align=center, minimum height=7mm,
              font=\scriptsize, inner sep=2pt},
  enc/.style={box, fill=black!4, minimum width=15mm},
  arr/.style={-{Latex[length=2mm]}, thin}]
\node[enc] (e1) {DINOv3\\ViT-H+};
\node[enc, below=of e1] (e2) {SigLIP2\\Giant};
\node[enc, below=of e2] (e3) {InternImage\\-H};
\node[box, right=of e2, minimum width=12mm] (dec) {SimpleFPN\\+ M2F};
\node[box, right=of dec, minimum width=14mm] (tta) {multi-scale\\+ flip TTA};
\node[box, right=of tta, minimum width=14mm, fill=black!8] (fuse) {per-class\\softmax\\fusion};
\foreach \x in {e1,e2,e3} \draw[arr] (\x.east) -- (dec.west);
\draw[arr] (dec) -- (tta);
\draw[arr] (tta) -- (fuse);
\node[below=1mm of e3, font=\scriptsize\itshape] {three diverse pretrainings};
\end{tikzpicture}
\caption{Overall architecture of our solution.
Three encoders with deliberately different pretraining share the same SimpleFPN and Mask2Former decoder. Multi-scale and flip test-time augmentation are leveraged after the output of each encoder. Their softmax outputs are fused with per-class validation-IoU weights.}
\label{fig:system}
\end{figure}
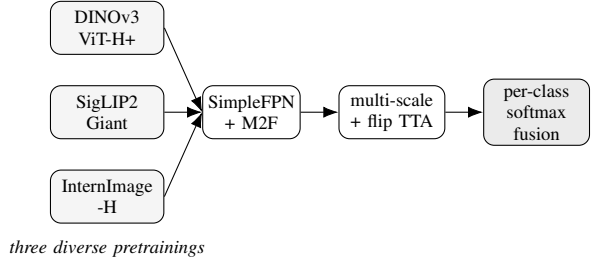

Every model in this report adopts the same architecture as illustrated in Fig.~\ref{fig:system}, which consists of a pretrained encoder, a SimpleFPN neck that maps four encoder taps to a common 256-channel pyramid, and a Mask2Former head with 100 queries.
For plain ViT encoders, we extract features from four evenly spaced blocks; for hierarchical encoders, we use features from the four native stages.
The decoder, query count, and loss are held fixed across all encoders so that accuracy differences can be primarily attributed to the encoder and its pretraining.

\subsection{Training recipe}
The backbone is fine-tuned end to end with a small backbone learning-rate multiplier.
We use AdamW with learning rate $10^{-4}$, backbone multiplier $0.01$, weight decay $0.05$, and gradient clipping with a max norm of $0.01$, with a linear warmup followed by polynomial decay.
Training augmentation is random resize with a scale ratio sampled from $[0.5, 2.0]$, a random crop, random horizontal flip, and photometric distortion.
Two design choices proved decisive once the encoder was chosen.
First, we enlarge the training crop from $512\times 512$ to $768\times 768$, which provides the visual encoder with a larger spatial context.
Second, we maintain an exponential moving average (EMA) of the weights and evaluate the averaged model, which stabilizes the long-tailed class learning over long schedules.
Training schedules range from 400k to 1M iterations with batch size set as 1.
Gradient checkpoint is adopted to save memory.

\subsection{Inference and test-time augmentation}
Because input resolutions vary by platform and exceed the training crop, we use sliding-window inference with a $768$ window and stride $512$.
At test time we average softmax outputs over three scales $\{0.75, 1.0, 1.25\}$ and a horizontal flip.
This test-time augmentation is the largest post-training gain we observe as shown in Section~\ref{sec:ablation}.

\subsection{Pretraining-diverse ensemble}
\label{sec:ensemble}
Our final submission ensembles three encoders chosen for \emph{different} pretraining objectives, that is, DINOv3 (self-supervised objective), SigLIP2 (vision-language contrastive objective), and InternImage (fully supervised objective).
The intuition is that different pretrained encoders produce diverse per-class errors, so a class-aware fusion can assign larger weights to the most reliable model for each category.
For model $m$ and class $c$ we set a fusion weight from each model's per-class validation IoU,
\begin{align}
w_{m,c} &= \frac{\mathrm{IoU}_{m,c}^{\,\alpha}}
               {\sum_{m'} \mathrm{IoU}_{m',c}^{\,\alpha}}, \\
p_c(x)  &= \sum_{m} w_{m,c}\,\mathrm{softmax}\big(z^{m}(x)\big)_c,
\end{align}
where $z^m$ is the logit map of model $m$ and $\alpha$ controls how sharply the weighting favors the locally best model ($\alpha\!=\!0$ recovers a uniform average).
We sweep $\alpha\in\{8,12,16\}$ on validation and keep the best.

\section{EXPERIMENTS}
\label{sec:ablation}

%
All models are trained in MMSegmentation~\cite{mmseg} on the combined GOOSE and GOOSE-Ex splits (11{,}234 training, 1{,}369 validation images).
Single-GPU training uses an RTX 4090 (24\,GB) for ViT-H+ and lighter encoders, with an A40 (46\,GB) or RTX 5880 (48\,GB) for the largest backbones.
Gradient checkpoint is adopted for experiments of ViT-H+ to enable $768\times 768$ input resolution.
The Mask2Former head uses the standard classification, mask, and dice losses with weights $2{:}5{:}5$ and a no-object weight of $0.1$.
Momentum of EMA is set to $10^{-4}$.
Reported validation numbers use the EMA weights and, where stated, test-time augmentation.

\subsection{Encoder pretraining recipe}
Table~\ref{tab:encoder-fixed} fixes the encoder scale (ViT-L) and the training settings (80k iterations, $512\times 512$ crop size) and varies only the pretraining objective.
At a single scale, the ordering tracks the objective: masked-image and self-supervised pretraining lead, contrastive vision-language pretraining follows, and an autoregressive objective trails.
Parameter count is constant across these rows, so the spread of 4.7 points is due to pretraining alone.

\begin{table}[t]
\centering
\caption{Different training objective of encoders (ViT-L, 80k iteration, $512\times 512$ crop size and Mask2Former decoder).
All rows have $\approx$300M backbone parameters.}
\label{tab:encoder-fixed}
\begin{tabular}{llc}
\toprule
Encoder & Pretraining objective & \miou \\
\midrule
AIMv2-L~\cite{aimv2}       & autoregressive          & 63.63 \\
MetaCLIP2-L~\cite{metaclip}& contrastive (CLIP)      & 64.43 \\
SigLIP2-L~\cite{siglip2}   & contrastive (sigmoid)   & 66.20 \\
EVA02-CLIP-L~\cite{eva02}  & contrastive (CLIP)      & 66.89 \\
DINOv3-L~\cite{dinov3}     & self-supervised         & 67.54 \\
EVA02-MIM-L~\cite{eva02}   & masked image modeling   & \textbf{68.34} \\
\bottomrule
\end{tabular}
\end{table}

Table~\ref{tab:encoder-recipe} then removes the budget confound: every encoder is trained under the same strong recipe ($768^2$, EMA, long schedule).
Here scale and architecture both vary, and the result is the core finding of this report.
The 3.0B-parameter SwinV2-Giant, an architecture-focused supervised model, reaches only 66.05\%, while the 840M-parameter self-supervised DINOv3 ViT-H+ reaches 74.68\%.
More tellingly, the two ConvNeXt rows hold the architecture fixed and change only the pretraining: a DINOv3-style distilled pretraining reaches 70.40\% (even before its schedule completes) against 68.64\% for masked autoencoding.
Parameter count does not order this table; pretraining does.

\begin{table}[t]
\centering
\caption{Encoder under a matched strong recipe ($768\times 768$ crop size, EMA, long schedule, and 
Mask2Former decoder). $^{\dagger}$schedule not yet complete.}
\label{tab:encoder-recipe}
\setlength{\tabcolsep}{4pt}
\begin{tabular}{llrc}
\toprule
Encoder & Pretraining & Params & \miou \\
\midrule
SwinV2-Giant~\cite{swinv2}        & IN-22k sup.\       & 3.0B   & 66.05 \\
ConvNeXtV2-H~\cite{convnextv2}    & masked AE          & 660M   & 68.64 \\
DINOv3-ConvNeXt-L~\cite{dinov3}   & self-sup.\ distill & 200M   & 70.40$^{\dagger}$ \\
InternImage-H~\cite{internimage}  & DCNv3, large-scale & 1.08B  & 72.64 \\
SigLIP2-Giant~\cite{siglip2}      & vision-language    & 1.87B  & 72.90 \\
DINOv3 ViT-H+~\cite{dinov3}       & self-supervised    & 840M   & \textbf{74.68} \\
\bottomrule
\end{tabular}
\end{table}

\subsection{Training and inference recipe}
Table~\ref{tab:recipe} decomposes the path from a baseline DINOv3 ViT-H+ to the final single-model configuration.
Enlarging the crop to $768\times 768$ adds 2.2 points; EMA adds about half a point; and multi-scale plus flip test-time augmentation adds a further 1.5 points, the largest single post-training gain.
Two results are negative and worth stating.
Replacing the Mask2Former head with a per-pixel UPerNet head costs 4.7 points, confirming that a mask-classification decoder is the right paradigm but is a settled choice rather than a tuning knob.
Removing photometric distortion slightly \emph{improves} the score, so the augmentation is at best neutral here.

\begin{table}[t]
\centering
\caption{Recipe and inference ablation on DINOv3 ViT-H+ (validation).
Each row changes one factor relative to the $768^2$+EMA model.}
\label{tab:recipe}
\begin{tabular}{lc}
\toprule
Configuration & \miou \\
\midrule
$512^2$ crop                          & 72.52 \\
$\;\;+\,768^2$ crop                   & 74.68 \\
$\;\;-\,$EMA                          & 74.10 \\
$\;\;-\,$photometric distortion       & 74.83 \\
UPerNet head (vs.\ Mask2Former)       & 70.03 \\
$\;\;+\,$multi-scale $+$ flip TTA     & \textbf{76.13} \\
\bottomrule
\end{tabular}
\end{table}

\subsection{The pretraining-diverse ensemble}
Table~\ref{tab:ensemble} reports the final ensemble results of our solution.
Each of the three encoders is individually strong (72.6--75.1\% on validation), and they were chosen for different pretraining objectives rather than peak single-model score.
The per-class weighted fusion of Section~\ref{sec:ensemble} edges out a uniform average.
On the held-out test set, the ensemble scores 75.40\% composite mIoU, against 75.12\% for the best single model and 29.22\% for the official baseline.

\begin{table}[t]
\centering
\caption{Final ensemble results. Test mIoU denotes the scores on the challenge leaderboard.}
\label{tab:ensemble}
\begin{tabular}{lcc}
\toprule
System & val \miou & test \miou \\
\midrule
DINOv3 ViT-H+ (best single)        & 75.14 & 75.12 \\
SigLIP2-Giant                      & 72.90 & --    \\
InternImage-H                      & 72.64 & --    \\
\midrule
Ensemble, per-class fusion (final) & --    & \textbf{75.40} \\
Official baseline~\cite{goose}     & --    & 29.22 \\
\bottomrule
\end{tabular}
\end{table}

The ensemble's test-set margin over the best single model is 0.28 point.
This result is consistent with our main observation: once a strong foundation encoder is well pretrained and sufficiently optimized, it already achieves very competitive performance on this benchmark. Pretraining-diverse ensembling therefore provides a small but consistent additional gain, rather than fundamentally changing the performance level.

\subsection{Strategies that did not transfer}
We also evaluated two color-normalization strategies emphasized by the 2025 challenge winner~\cite{goose2025_2d}.
Color-shift correction applied to the training and validation lists reduced composite mIoU by 0.8 point relative to the matched baseline, and photometric distortion was neutral to slightly negative (Table~\ref{tab:recipe}).
We attribute this to the encoders themselves: a backbone pretrained on billions of images has already seen wide exposure and white-balance variation, so explicit cross-platform color correction is redundant and occasionally harmful.


\section{CONCLUSION}
For fine-grained off-road semantic segmentation on GOOSE, the choice that matters most is how the encoder was pretrained.
Under a fixed decoder and budget, foundation-model encoders outperform architecture-focused ones by margins that parameter count cannot explain, and the largest model we tried is the weakest.
Once a foundation encoder is in hand, a plain recipe of long training, weight averaging, a larger crop, and multi-scale plus flip test-time augmentation carries a single model to 76\% composite mIoU, and a pretraining-diverse ensemble adds a final increment to a 75.40\% test result.
We hope the controlled comparison is useful to others building outdoor robotic perception systems.

\bibliographystyle{IEEEtran}
\bibliography{references}

@inproceedings{goose,
  title={The goose dataset for perception in unstructured environments},
  author={Mortimer, Peter and Hagmanns, Raphael and Granero, Miguel and Luettel, Thorsten and Petereit, Janko and Wuensche, Hans-Joachim},
  booktitle={2024 IEEE International Conference on Robotics and Automation (ICRA)},
  pages={14838--14844},
  year={2024},
  organization={IEEE}
}

@article{goose2025_2d,
  title={Technical Report for ICRA 2025 GOOSE 2D Semantic Segmentation Challenge: Leveraging Color Shift Correction, RoPE-Swin Backbone, and Quantile-based Label Denoising Strategy for Robust Outdoor Scene Understanding},
  author={Hsu, Chih-Chung and Wu, I and Tseng, Wen-Hai and Cheng, Ching-Heng and Wu, Ming-Hsuan and Jiang, Jin-Hui and Hsiao, Yu-Jou and others},
  journal={arXiv preprint arXiv:2505.06991},
  year={2025}
}

@inproceedings{mask2former,
  title={Masked-attention mask transformer for universal image segmentation},
  author={Cheng, Bowen and Misra, Ishan and Schwing, Alexander G and Kirillov, Alexander and Girdhar, Rohit},
  booktitle={Proceedings of the IEEE/CVF conference on computer vision and pattern recognition},
  pages={1290--1299},
  year={2022}
}

@inproceedings{upernet,
  title={Unified perceptual parsing for scene understanding},
  author={Xiao, Tete and Liu, Yingcheng and Zhou, Bolei and Jiang, Yuning and Sun, Jian},
  booktitle={Proceedings of the European conference on computer vision (ECCV)},
  pages={418--434},
  year={2018}
}

@inproceedings{segformer,
  title={SegFormer: Simple and efficient design for semantic segmentation with transformers},
  author={Xie, Enze and Wang, Wenhai and Yu, Zhiding and Anandkumar, Anima and Alvarez, Jose M and Luo, Ping},
  journal={Advances in neural information processing systems},
  volume={34},
  pages={12077--12090},
  year={2021}
}

@inproceedings{swin,
  title={Swin transformer: Hierarchical vision transformer using shifted windows},
  author={Liu, Ze and Lin, Yutong and Cao, Yue and Hu, Han and Wei, Yixuan and Zhang, Zheng and Lin, Stephen and Guo, Baining},
  booktitle={Proceedings of the IEEE/CVF international conference on computer vision},
  pages={10012--10022},
  year={2021}
}

@inproceedings{swinv2,
  title={Swin transformer v2: Scaling up capacity and resolution},
  author={Liu, Ze and Hu, Han and Lin, Yutong and Yao, Zhuliang and Xie, Zhenda and Wei, Yixuan and Ning, Jia and Cao, Yue and Zhang, Zheng and Dong, Li and others},
  booktitle={Proceedings of the IEEE/CVF conference on computer vision and pattern recognition},
  pages={12009--12019},
  year={2022}
}

@inproceedings{convnext,
  title={A convnet for the 2020s},
  author={Liu, Zhuang and Mao, Hanzi and Wu, Chao-Yuan and Feichtenhofer, Christoph and Darrell, Trevor and Xie, Saining},
  booktitle={Proceedings of the IEEE/CVF conference on computer vision and pattern recognition},
  pages={11976--11986},
  year={2022}
}

@inproceedings{convnextv2,
  title={Convnext v2: Co-designing and scaling convnets with masked autoencoders},
  author={Woo, Sanghyun and Debnath, Shoubhik and Hu, Ronghang and Chen, Xinlei and Liu, Zhuang and Kweon, In So and Xie, Saining},
  booktitle={Proceedings of the IEEE/CVF conference on computer vision and pattern recognition},
  pages={16133--16142},
  year={2023}
}

@article{dinov3,
  title={Dinov3},
  author={Sim{\'e}oni, Oriane and Vo, Huy V and Seitzer, Maximilian and Baldassarre, Federico and Oquab, Maxime and Jose, Cijo and Khalidov, Vasil and Szafraniec, Marc and Yi, Seungeun and Ramamonjisoa, Micha{\"e}l and others},
  journal={arXiv preprint arXiv:2508.10104},
  year={2025}
}

@article{siglip2,
  title={Siglip 2: Multilingual vision-language encoders with improved semantic understanding, localization, and dense features},
  author={Tschannen, Michael and Gritsenko, Alexey and Wang, Xiao and Naeem, Muhammad Ferjad and Alabdulmohsin, Ibrahim and Parthasarathy, Nikhil and Evans, Talfan and Beyer, Lucas and Xia, Ye and Mustafa, Basil and others},
  journal={arXiv preprint arXiv:2502.14786},
  year={2025}
}

@inproceedings{internimage,
  title={Internimage: Exploring large-scale vision foundation models with deformable convolutions},
  author={Wang, Wenhai and Dai, Jifeng and Chen, Zhe and Huang, Zhenhang and Li, Zhiqi and Zhu, Xizhou and Hu, Xiaowei and Lu, Tong and Lu, Lewei and Li, Hongsheng and others},
  booktitle={Proceedings of the IEEE/CVF conference on computer vision and pattern recognition},
  pages={14408--14419},
  year={2023}
}

@inproceedings{eva02,
  title={Eva-02: A visual representation for neon genesis},
  author={Fang, Yuxin and Sun, Quan and Wang, Xinggang and Huang, Tiejun and Wang, Xinlong and Cao, Yue},
  journal={Image and Vision Computing},
  volume={149},
  pages={105171},
  year={2024},
  publisher={Elsevier}
}

@article{aimv2,
  title={Multimodal autoregressive pre-training of large vision encoders},
  author={Fini, Enrico and Shukor, Mustafa and Li, Xiujun and Dufter, Philipp and Klein, Michal and Haldimann, David and Aitharaju, Sai and da Costa, Victor G Turrisi and B{\'e}thune, Louis and Gan, Zhe and others},
  booktitle={Proceedings of the IEEE/CVF Conference on Computer Vision and Pattern Recognition},
  pages={9641--9654},
  year={2025}
}

@inproceedings{metaclip,
  title={Demystifying clip data},
  author={Xu, Hu and Xie, Saining and Tan, Xiaoqing and Huang, Po-Yao and Howes, Russell and Sharma, Vasu and Li, Shang-Wen and Ghosh, Gargi and Zettlemoyer, Luke and Feichtenhofer, Christoph},
  booktitle={International Conference on Learning Representations},
  volume={2024},
  pages={47812--47831},
  year={2024}
}

@misc{mmseg,
  title        = {MMSegmentation: Openmmlab semantic segmentation toolbox and benchmark},
  author       = {{MMSegmentation Contributors}},
  howpublished = {\url{https://github.com/open-mmlab/mmsegmentation}},
  year         = {2020}
}

@article{huang2020ordnet,
  title={ORDNet: Capturing omni-range dependencies for scene parsing},
  author={Huang, Shaofei and Liu, Si and Hui, Tianrui and Han, Jizhong and Li, Bo and Feng, Jiashi and Yan, Shuicheng},
  journal={IEEE Transactions on Image Processing},
  volume={29},
  pages={8251--8263},
  year={2020},
  publisher={IEEE}
}

@article{huang2024modality,
  title={Modality adaptation via feature difference learning for depth human parsing},
  author={Huang, Shaofei and Hui, Tianrui and Gong, Yue and Peng, Fengguang and Fang, Yuqiang and Wang, Jingwei and Ma, Bin and Wei, Xiaoming and Han, Jizhong},
  journal={Computer Vision and Image Understanding},
  volume={247},
  pages={104070},
  year={2024},
  publisher={Elsevier}
}

@inproceedings{wang2023transferring,
  title={Transferring CLIP's knowledge into zero-shot point cloud semantic segmentation},
  author={Wang, Yuanbin and Huang, Shaofei and Gao, Yulu and Wang, Zhen and Wang, Rui and Sheng, Kehua and Zhang, Bo and Liu, Si},
  booktitle={Proceedings of the 31st ACM International Conference on Multimedia},
  pages={3745--3754},
  year={2023}
}

@inproceedings{huang2025revisiting,
  title={Revisiting audio-visual segmentation with vision-centric transformer},
  author={Huang, Shaofei and Ling, Rui and Hui, Tianrui and Li, Hongyu and Zhou, Xu and Zhang, Shifeng and Liu, Si and Hong, Richang and Wang, Meng},
  booktitle={Proceedings of the Computer Vision and Pattern Recognition Conference},
  pages={8352--8361},
  year={2025}
}

@inproceedings{fpn,
  title={Fully convolutional networks for semantic segmentation},
  author={Long, Jonathan and Shelhamer, Evan and Darrell, Trevor},
  booktitle={Proceedings of the IEEE conference on computer vision and pattern recognition},
  pages={3431--3440},
  year={2015}
}

\end{document}